\documentclass[letterpaper, 10 pt, conference]{ieeeconf}
\IEEEoverridecommandlockouts
\usepackage{cite}
\usepackage{amsmath,amssymb,amsfonts}
\usepackage{algorithmic}
\usepackage{graphicx}
\usepackage{textcomp}
\usepackage{xcolor}
\usepackage{makecell}
\usepackage{bm}
\usepackage{subfigure}
\usepackage{wasysym}
\def\BibTeX{{\rm B\kern-.05em{\sc i\kern-.025em b}\kern-.08em
    T\kern-.1667em\lower.7ex\hbox{E}\kern-.125emX}}
\begin{document}

\title{\LARGE \bf
Gravity-aware Grasp Generation with Implicit Grasp Mode Selection\\for Underactuated Hands
}

\author{Tianyi Ko$^{*\dagger}$, Takuya Ikeda$^*$, Thomas Stewart$^*$, Robert Lee$^*$, Koichi Nishiwaki$^*$
\thanks{$*$ T. Ko, T. Ikeda, T. Stewart, R. Lee, K. Nishiwaki are with Woven by Toyota, Inc., 3-2-1 Nihonbashi-Muromachi, Chuo-ku, Tokyo, Japan.}
\thanks{$\dagger$ Corresponding author. \texttt{tianyi.ko@woven.toyota}}
}

\maketitle
\thispagestyle{empty}
\pagestyle{empty}

\begin{abstract}
Learning-based grasp detectors typically assume a precision grasp, where each finger only has one contact point, and estimate the grasp probability.
In this work, we propose a data generation and learning pipeline that can leverage power grasping, which has more contact points with an enveloping configuration and is robust against both positioning error and force disturbance.
To train a grasp detector to prioritize power grasping while still keeping precision grasping as the secondary choice, we propose to train the network against the magnitude of disturbance in the gravity direction a grasp can resist (gravity-rejection score) rather than the binary classification of success.
We also provide an efficient data generation pipeline for a dataset with gravity-rejection score annotation.
Evaluation in both simulation and real-robot clarifies the significant improvement in our approach, especially when the objects are heavy.
\end{abstract}

\section{Introduction}
\label{sec:introduction}
The majority of state-of-the-art machine learning-based grasp detectors \cite{s4g, regnet, contact_graspnet, vgn, giga, gsnet, edge_grasp, vpn} assume that each rigid finger only has one contact with a small contact region.
Although such a grasp is easy to handle in both data generation and learning aspects, its limited contact region makes the grasp fragile.
A typical failure mode with such grasp mode by a parallel-jaw gripper is that the object rotates around the axis connecting the two contact points, during which a translational displacement also occurs, resulting in the object dropping.
The small contact region also imposes a high demand for accurate hand-eye calibration and hand positioning, which is challenging for low-cost manipulators or those under low-impedance control for safety.

Napier~\cite{napier1956prehensile} pointed out that in order to firmly grasp an object, humans utilize the palm, and referred to such grasps as a \textit{power grip.}
Cutkosky~\cite{cutkosky1989grasp} further extended the taxonomy of \cite{napier1956prehensile} in the robotics context and defined \textit{power grasp} as grasps distinguished by large areas of contact between the grasped object and the surfaces of the fingers and palm, while classifying grasps where the object is held with the tips of the fingers and thumb
as \textit{precision grasp.}
Despite the fact that there are multiple definitions of \textit{power grasp}, e.g., Zhang et al.~\cite{zhang1994robustness} redefined it as a type of grasp that its mechanism can resist passively against external forces without relying on feedback control of joint torques, and Zhang et al.~\cite{zhang1995definition} redefined it as a grasp with zero or less than zero connectivity, they share the same idea that utilizing multiple contacts of the intermediate finger links and palm improves grasp quality.
In this paper, we introduce a data generation and learning framework to learn \textit{power grasps} (with the definition of \cite{cutkosky1989grasp}) in addition to the common \textit{precision grasps}, as shown in Figure~\ref{fig:gravity_projection} (a).

\begin{figure}[tb]
    \centerline{\includegraphics[width=0.99\columnwidth]{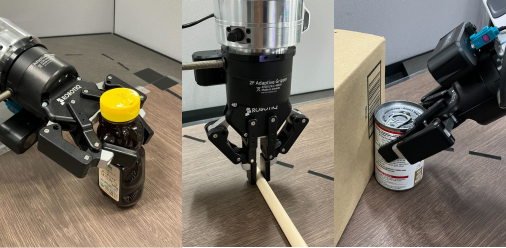}}
    \caption{
    While existing works only handle precision grasping, power grasping is more robust against both initial position error and post-grasp force disturbance.
    We propose a data generation pipeline and neural network model that prioritizes power grasping if applicable (left) but implicitly switches back to precision grasping if power grasping is not available due to collision with the table (middle) or other objects (right.)
    }
    \label{fig:grasp_pose}
\end{figure}

Underactuated hands~\cite{handbook_hand} have a larger number of joints than actuators.
This passive compliance allows the hand to adapt to various objects with non-typical geometries without explicitly controlling each joint.
Since it makes power grasping easier, this work focuses on utilizing underactuated hands, or more specifically, Robotiq 2F-85~\cite{robotiq}.
However, such mechanically automatic property makes learning power grasping even challenging because whether the grasp is a power grasp or a precision grasp cannot be explicitly controlled but only decided through the interaction between the object and the hand's link mechanism.
In addition, since a power grasp requires the fingers to wrap around the object, which is not always possible due to collisions with the environment, such as the tabletop or other objects (see Fig.~\ref{fig:grasp_pose} middle and right.)
To this end, we propose a neural network architecture that prioritizes power grasping if applicable but implicitly switches back to precision grasping if power grasping is not available.
More specifically, while prior works~\cite{s4g, regnet, contact_graspnet, vgn, giga, gsnet, edge_grasp, vpn} only estimate the success probability of a grasp, we handle two values, namely (i) \textit{gravity-rejection score} that represents the magnitude of disturbance in the gravity direction a grasp can support, which implicitly encourages power grasping since our data generation pipeline labels higher score for power grasps, and (ii) \textit{grasp validness} which implicitly rejects grasps where the fingers will collide with the environment or it is too big for the hand to grasp.

Our contributions are as follows:
\begin{itemize}
    \item We propose to measure the grasp quality by \textit{gravity-rejection score}, which is the magnitude of disturbance in the gravity direction a grasp can support in the simulation.
    \item We provide a data generation / learning pipeline that can generate power grasps while automatically switching to precision grasping when power grasping is not available due to collision.
    \item We propose to evaluate a grasp detector with varying object weight in order to examine both the detection accuracy and robustness of the detected grasp.
\end{itemize}

\section{Related Works}
\subsection{Training Data for Learned Grasp Detectors}
\label{sec:related_data}
Grasp quality metrics are key in generating synthetic training data for learning grasp detectors.
An often selected approach is based on an antipodal grasp~\cite{antipodal}, adopted in \cite{s4g, regnet, vpn}.
By definition, though, it only supports antipodal grasps and is thus not applicable to power grasps with more than two contact points.
Ferrari and Canny~\cite{optimal_grasps} proposed Q1 criteria, which is the maximum radius of a $\mathbb{R}^6$ sphere in the contact wrench space (GWS), with its center aligned with the origin, under the constraint that the L1 norm of the contact forces is one.
Frequently referred to as $\epsilon$-metrics, it is adopted in many works such as \cite{dexnet2, graspnet-1billion, gsnet, dlr_autoregression, vpn, anygrasp}.
Weisz and Allen~\cite{weisz2012pose} discussed the robustness of the $\epsilon$-metrics against the grasp pose error.
As the $\epsilon$-metrics' assumption that each contact can exert arbitrary reaction force within the friction cone limited its application, Winkelbauer et al.~\cite{dlr_autoregression} extended $\epsilon$-metrics to consider the torque constraint of articulated fingers.
Nevertheless, the constraint of the $\epsilon$-metrics on the reaction force L1 norm prevents it from being adopted for power grasps that utilize kinematic constraint forces.
To provide a quality metric for power grasps, Mirzal et al.~\cite{mirza1994_general} discussed the stability region of a power grasp.
Zhang et al.~\cite{zhang1994robustness} proposed a virtual work-based quality measure.
However, although these works assume fully actuated hands, many robot hands that support power grasping (including ours) have an underactuated mechanism, making it hard to apply those metrics directly.

While analytical metrics are computationally efficient, it is difficult to capture all aspects of a contact, such as contact force distribution or motion after a slip.
Kappler et al.~\cite{leveraging_big_data} reported that a simulation-based approach achieves a better performance.
Zhou et al.~\cite{6dof_grasp_planning} and Eppner et al.~\cite{acronym} followed \cite{leveraging_big_data} by shaking the hand in a gravity-less simulation.
Breyer et al.~\cite{vgn} and Jiang et al.~\cite{giga} evaluated grasps by actually lifting the object in a simulation and classified its success or failure.
Huang et al.~\cite{edge_grasp} further shake the hand after lifting the object to select only robust grasps.
Those prior works use the simulator to get a set of valid grasps.
However, such binary classification suffers from the sim-to-real gap since the absolute value of success or failure highly depends on the simulator's contact model and parameters.
In this paper, we also take a simulation-based approach, but we acquire a continuous spectrum of the \textit{gravity-rejection score} to train the neural network to prioritize grasps with higher safety margins.
A similar idea can be found in Fang et al.~\cite{anygrasp}, where a score based on the distance from the object's center of mass was used, but our approach considers more diverse cases such as power grasping.
In addition, we propose an approximation to allow the two-staged approach proposed by Qin et al.~\cite{s4g} to relax the high computation cost of the simulation-based approaches.
Section~\ref{sec:datagen} details our data generation pipeline.

\subsection{Grasp Representation in Learned Graps Detectors}
\label{sec:related_network}
\begin{figure}[tbp]
    \centerline{
    \subfigure[]{\includegraphics[width=0.5\linewidth]{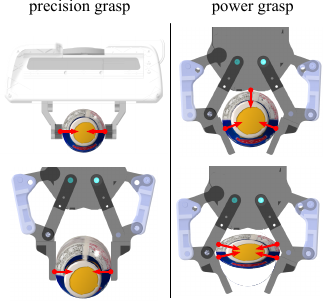}}
    \subfigure[]{\includegraphics[width=0.4\linewidth]{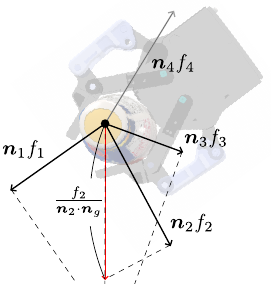}}
    }
    \caption{
    (a) Most works assume a precision grasp (left) due to its simplicity. A power grasp (right) allows a more robust grasp thanks to the larger number of contacts.
    (b) Illustration of Eq.~\ref{eq:gravity_projection} which approximates the multidimensional \textit{disturbance-rejection score} to the gravity direction \textit{gravity-rejection score}.
    }
    \label{fig:gravity_projection}
\end{figure}

Grasp representation is another key to learning grasp detectors.
For 2D top-down grasps, the grasps were typically represented as the hand's x-y location and yaw rotation on the height map~\cite{dexnet2,ggcnn}.
This approach is difficult to apply to power grasping because the hand's insertion depth cannot be heuristically derived.
For grasp detection in 3D space, Ten Pas et al.~\cite{gpd_ijrr} first sampled points on the object surface and then searched for grasp candidates in the Darboux frame, followed by a learned grasp classifier. However, their ``push" operation makes it difficult to detect both power grasps and precision grasps simultaneously.
Qin et al.~\cite{s4g} regressed translation offset and rotation of the hand from each point on the object surface.
Cai et al.~\cite{vpn} used the location and normal of the points on the object surface to decide five degrees of freedom of the hand and regressed the remaining rotation in the plane.
Since \cite{s4g, vpn} interpreted the surface point as candidates facing the hand, a set of precision grasp and power grasp may be assigned to a single point, leading to representation ambiguity.
Zhao et al.~\cite{regnet} detected the middle point of two contact points to reconstruct a grasp, thus only supporting antipodal grasps.
Sundermeyer et al.~\cite{contact_graspnet} and Huang et al.~\cite{edge_grasp} interpreted surface points as candidates of contact and reported superior performance.
However, contact points of a power grasp significantly differ depending on object size, making explicit contact representation challenging (see Fig.~\ref{fig:gravity_projection} (a) where the contact points differ among different grasp modes.)

Wang et al.~\cite{gsnet} first detected the ``view" and evaluated grasps represented in the ``view" coordinate with discretized hand depth and in-plane rotation.
This approach is applicable to power grasping because a precision grasp and a power grasp are assigned to different discretized hand depths (see Fig.~\ref{fig:gravity_projection} (a) where the distance from the object differs for the different grasp modes).
Breyer et al.~\cite{vgn} and Jiang et al.~\cite{giga} voxelized the workspace and detected per-voxel grasp probability and hand rotation.
This representation is also applicable to detecting power grasps because (i) a precision grasp and a power grasp are assigned to different voxels, (ii) a precision grasp and a power grasp that grasps a similar location on the object tend to have a similar rotation so that we can expect a continuous grasp rotation field.
In this work, we follow this voxel approach due to the simple architecture and match for power grasping, but train the network with the \textit{gravity-rejection score}. 

\section{Synthetic Data Generation with Gravity-Rejection Score}
\label{sec:datagen}

\begin{figure}[tbp]
    \centerline{\includegraphics[width=0.72\columnwidth]{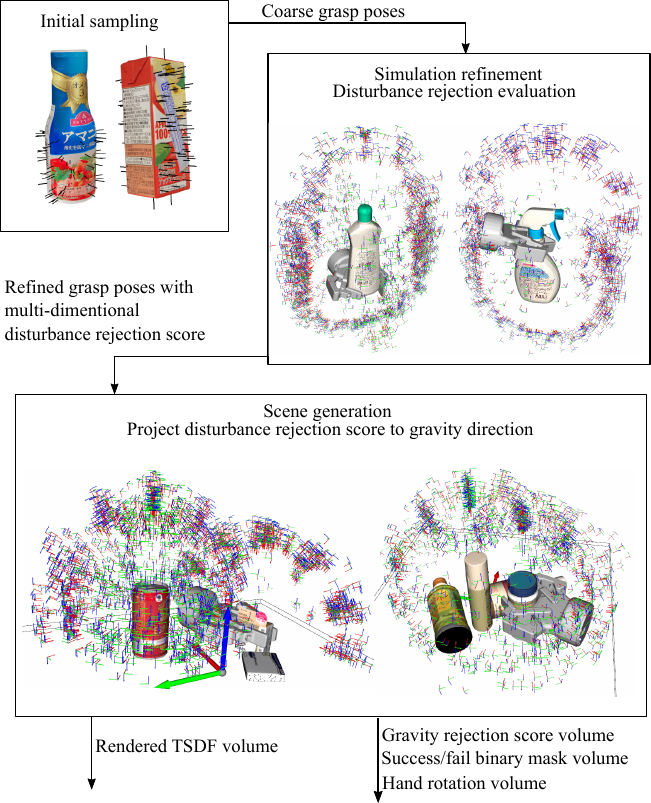}}
    \caption{
    Schematic of our data generation pipeline.
    We first sample antipodal grasps and randomize them (top-left).
    Those initial grasps are refined through grasp simulation (top-right).
    We apply external force in multiple directions and acquire a multi-dimensional \textit{disturbance-rejection score} for each grasp.
    Finally, we create multi-object scenes and project per-object grasp poses to the scene to acquire the training data (bottom).
    The \textit{disturbance-rejection score} is projected to the scene's gravity direction to acquire the \textit{gravity-rejection score}.
    }`
    \label{fig:data_generation}
\end{figure}

Given a set of scenes (in this work, 5K) with multiple objects (in this work, 1-5) placed in a cluttered arrangement (see Fig.~\ref{fig:data_generation} bottom), our target is to generate a dense annotation (in this work, 1.2K on average for each scene) of grasps with both power grasp and precision grasp mode, and label the \textit{gravity-rejection score}, which stands for the magnitude of disturbance in the gravity direction that the grasp can support.

Following the previous works~\cite{s4g, regnet, vpn}, we first generate per-object grasp poses.
We first generate precision grasps by sampling antipodal points on the object surface.
In order to generate power grasps, we perturb precision grasps in translation and rotation and simulate the contact process by closing the hand against the object in a gravityless simulation.
Through the contact, the object settles in the hand, and we then record the hand's pose relative to the object.
This simulation-based approach is beneficial because we can avoid explicitly modeling the complex mechanism of underactuated fingers, where the final hand configuration is only decided through the interaction between the object and the spring-loaded closed-link mechanism.
For each scene, we project the per-object grasp poses to the scene based on the object's pose, and remove those that collide with the table or other objects.

In order to train a network to prioritize power grasping, we need a grasp metric that takes on a higher value for power grasps.
As discussed in Sec.~\ref{sec:related_data}, we take a simulation-based approach rather than an analytical one~\cite{antipodal, optimal_grasps, mirza1994_general, zhang1994robustness}.
However, unlike the case of \cite{vgn, giga, edge_grasp} where the grasps are classified as success or failure, we measure the maximum disturance in the gravity direction that a grasp can support in the simulation.
We call this a \textit{gravity-rejection score}, which has a continuous value with the unit of [N].

One drawback of the simulation-based approach is the computation time.
In our case, the naive approach where each grasp in each scene is evaluated by a simulation will result in roughly 6M simulations with varying gravity.
While this is not intractable, thanks to modern cloud parallel computing, this paper proposes an approximation to reduce the number of simulations.
Specifically, we perform the simulation in the per-object stage rather than the per-scene stage.
For each grasp, in a gravity-free simulation, we apply an increasing external force to the object until it leaves the hand and then record the magnitude of that force.
By applying this force to the object's center of mass, we can encourage the network to prioritize grasping near the center of mass, in addition to prioritizing power grasping.

Since the object's pose relative to the scene's gravity direction is unknown, we perform this evaluation in multiple directions, resulting in an $n$-dimentional \textit{disturbance-rejection score} where $n$ is the number of directions (in this work, 6).
In order to project the $n$-dimentional \textit{disturbance-rejection score} to the scene's gravity direction to acquire the one-dimensional \textit{gravity-rejection score} $f_g$, we take the following approximation:
\begin{equation}
    f_g = \underset{\{ i | \bm{n}_i \cdot \bm{n}_g > \epsilon\}}\min\frac{f_i}{\bm{n}_i \cdot \bm{n}_g}
\label{eq:gravity_projection}
\end{equation}
where $\bm{n}_g \in \mathbb{R}^3$ is the unit vector heading the direction of gravity in a scene, $f_i$ is the $i$-th element of the \textit{disturbance-rejection score}, $\bm{n}_i  \in \mathbb{R}^3$ is the unit vector heading the direction of the $i$-th disturbance projected to the scene frame, and $\epsilon$ is a small positive value.
Figure~\ref{fig:gravity_projection} (b) illustrates the approximation and Fig.~\ref{fig:data_generation} illustrates the whole data generation pipeline.

Thanks to the reduction in the number of simulations, we can employ accurate yet time-consuming simulation setups.
Specifically, we use the hydroelastic contact model~\cite{hydroelastic_2019} implemented in Drake~\cite{drake}, which is highly realistic even with a concave-shaped object mesh.
Simulation for all grasps for a single object typically takes roughly 12-18 hours with a single AWS m5.2xlarge instance, but it is easy to parallelize since each simulation is independent.
In addition, the computation time highly depends on the settings.
For example, a coarse mesh speeds up the simulation in exchange for a less accurate object shape, a larger simulation step linearly improves the simulation time in exchange for a higher chance of failure, and simply switching the contact model to the point-contact results in a more than x10 faster but much more unstable simulation.

\section{Network Architecture and Training}
\label{sec:network}
Given a single depth image as the input, our target is to infer a set of grasp poses in SE(3) with \textit{gravity-rejection score}.
At the inference time, the robot tries to execute the first reachable and collision-free grasp with the highest \textit{gravity-rejection score}.
This differs from other works that estimate the probability of the grasp: a higher \textit{probability} means the network is more \textit{confident}, but it doesn't mean the grasp is more \textit{robust}; in our case the \textit{gravity-rejection score} has a unit of [N], which directly represents the grasp's robustness in the gravity direction.

As discussed in Section~\ref{sec:related_network}, we take a 3D fully convolutional network as the backbone following \cite{vgn}, but with a \textit{gravity-rejection score} regression head.
Unlike the case of \cite{vgn}, where both positive and negative samples were included in the dataset, our data generation pipeline described in Sec.~\ref{sec:datagen} only provides positive samples.
We therefore train the \textit{gravity-rejection score} head with L2 loss against positive examples only, leaving the invalid grasp regions as out-of-domain.
While \cite{vgn} introduced a heuristic filter to remove out-of-domain voxels, in this work, we propose to learn a \textit{grasp validness} head to classify whether the voxel is out-of-domain.
The evaluation in Sec.~\ref{sec:qualitative} shows that this classification head is also effective in collision avoidance or rejecting grasping an improper region.
As the training data is densely annotated, we fill all voxels without a grasp label as \textit{invalid} and use a weighted cross-entropy loss for the training.

\section{Experiment and Evaluation}
\subsection{Analysis of Gravity-Rejection Score and Classification}
\label{sec:qualitative}
\begin{figure}[tp]
    \centerline{\includegraphics[width=0.8\columnwidth]{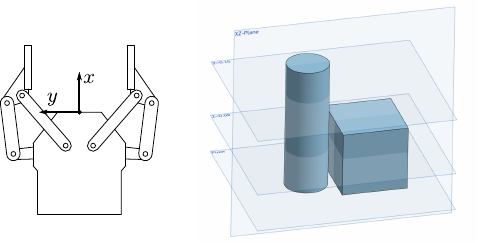}}
    \caption{
    TCP coordinate definition (left) and scene for the static network input/output analysis (right.)
    A $\diameter$65 $\times$ 200 mm cylinder and a 100 $\times$ 80 $\times$ 100 mm cube are placed on the surface and the input/output of the network is visualized with the three blue-colored cross-sections.
    }
    \label{fig:volume_experiment}
\end{figure}

\begin{figure}[tp]
    \centerline{\includegraphics[width=0.99\columnwidth]{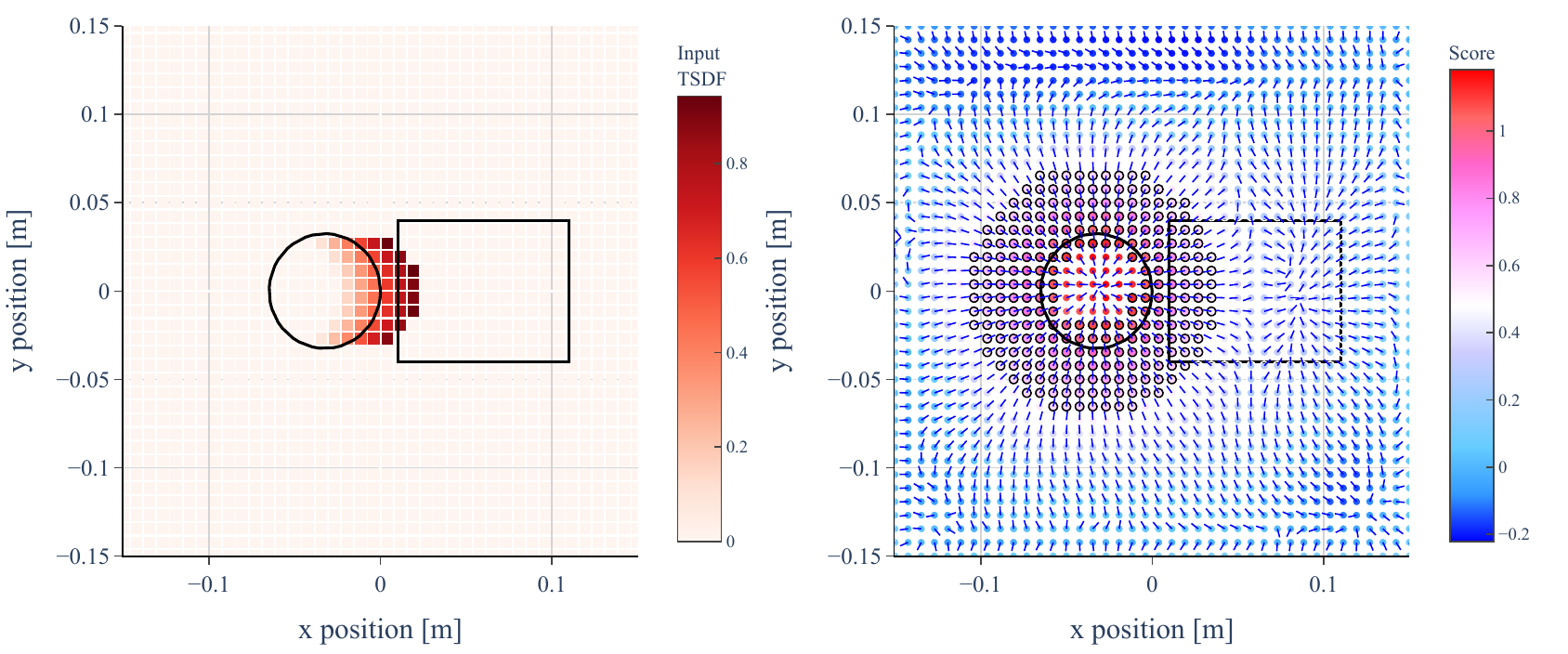}}
    \caption{
    Input TSDF volume (left) and output grasp volume (right) with $z=0.15$ cross-section.
    The markers with black-circle are voxels with positive classification.
    The short blue lines are the hand approach direction.
    }
    \label{fig:volume_horizontal_15}
\end{figure}

\begin{figure}[tp]
    \centerline{\includegraphics[width=0.99\columnwidth]{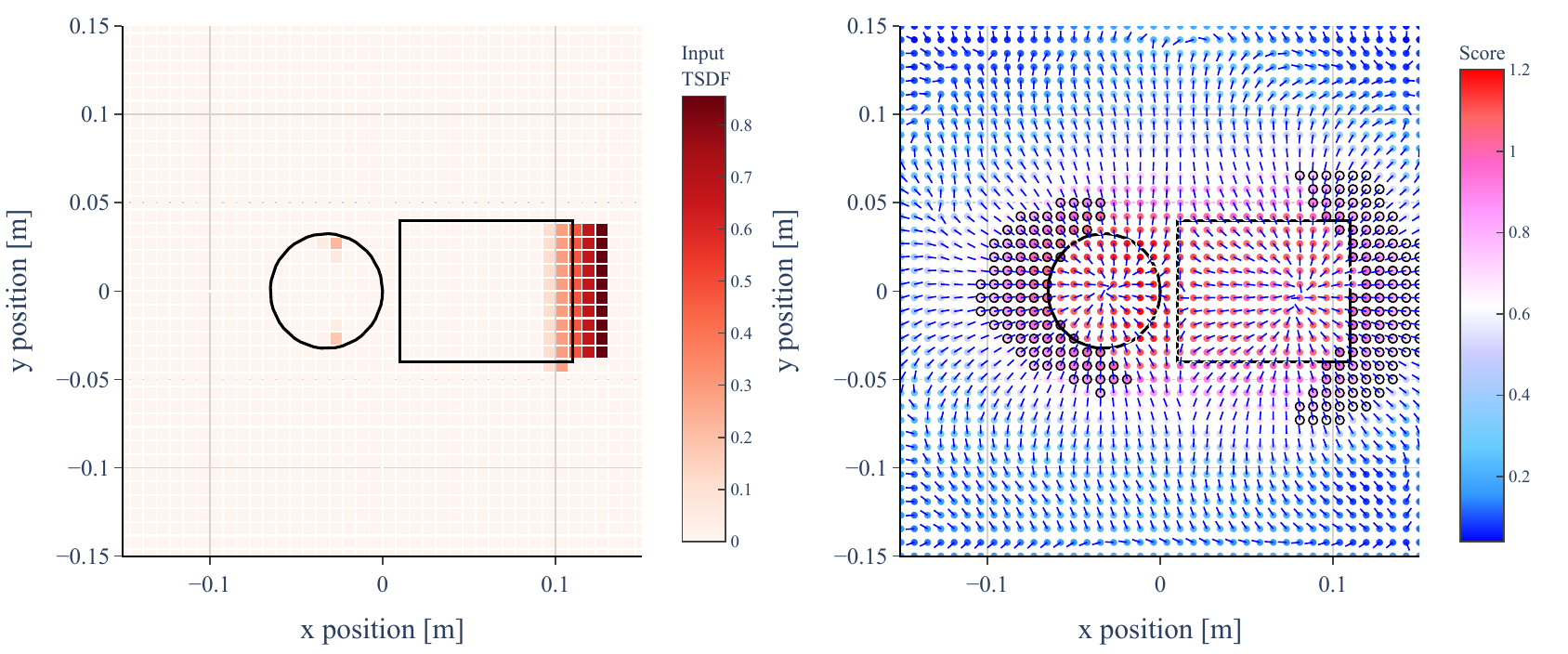}}
    \caption{
        Input TSDF volume (left) and output grasp volume (right) with $z=0.08$ cross-section.
        The score prioritizes power grasps while the classification rejects grasps with collision or with too-big width.
    }
    \label{fig:volume_horizontal_8}
\end{figure}

\begin{figure}[tp]
    \centerline{\includegraphics[width=0.99\columnwidth]{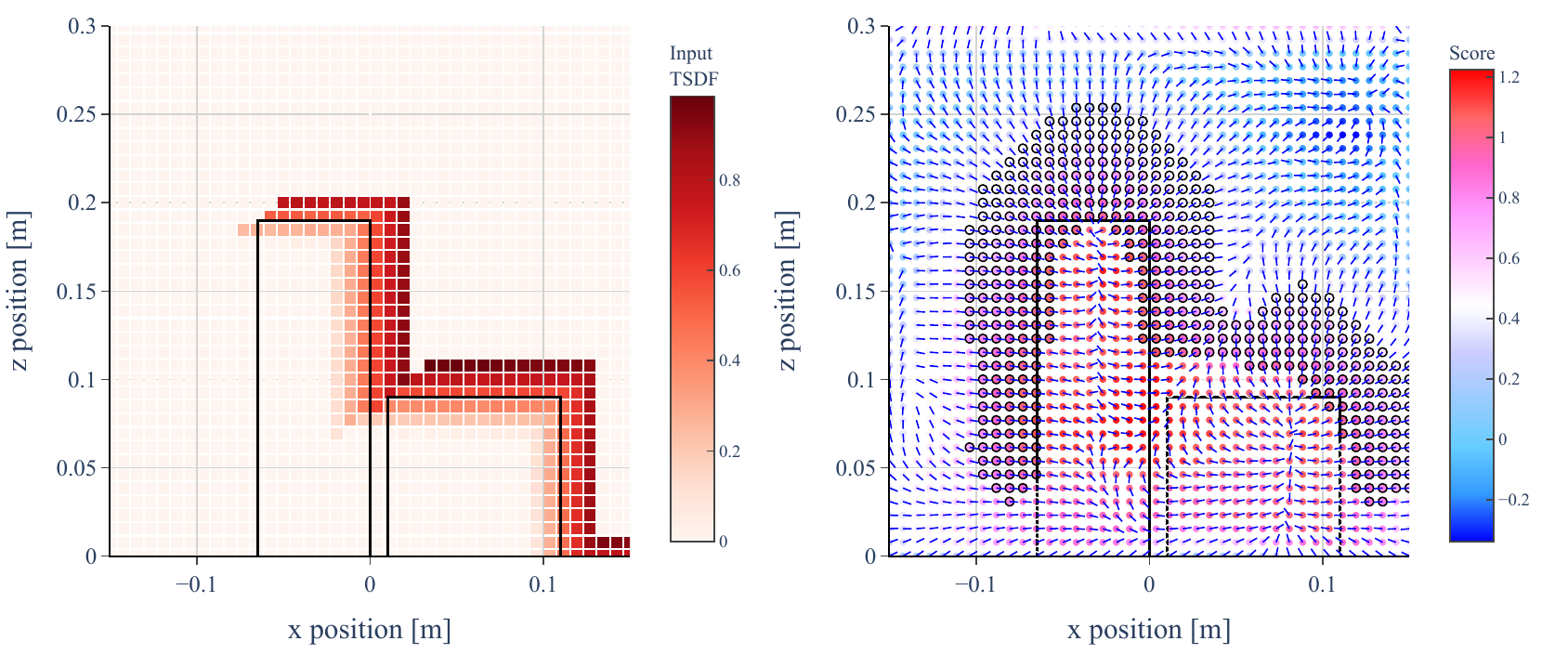}}
    \caption{
    Input TSDF volume (left) and output grasp volume (right) with $y=0$ cross-section.
    Even for the region invisible to the camera, the network well predicts grasp poses.
    The classification properly removes grasps where the palm may collide with other objects.
    }
    \label{fig:volume}
\end{figure}

This subsection provides a qualitative analysis of the network output \textit{gravity-rejection score} and \textit{grasp validness}.
We placed two primitive objects in the simulator: a $\diameter$65 $\times$ 200 mm cylinder at the location $[-0.0325, 0, 0]$ m and a 100$\times$80$\times$100 mm cube at $[0.06, 0, 0]$ m.
Even though these two objects were not included in our dataset, their geometries fall into our typical cases, such as water bottles and daily goods boxes, because we target home-use products.
We placed a single camera at $[0.5, 0, 0.5]$ [m] and plotted the input TSDF volume and the output grasp volume.
Figure~\ref{fig:volume_experiment} left illustrates the TCP (tool center point) coordinate, and Fig.~\ref{fig:volume_experiment} right shows the experiment scene.
We plot the cross-sections by $z=0.15$, $z=0.08$ and $y=0$ plane in Fig.~\ref{fig:volume_horizontal_15}, \ref{fig:volume_horizontal_8}, \ref{fig:volume} respectively.

The color bar on the right plot of the figures stands for the \textit{gravity-rejection score}, which is a continuous field that takes on a higher value near the object.
This distribution encourages power grasping, as shown in Fig.~\ref{fig:gravity_projection} (a).
Since the \textit{gravity-rejection score} head is only trained against valid grasps, its output for the out-of-domain voxels is incorrect, e.g., it takes on an even higher value inside the object and a medium score at regions far removed from the object.
The markers surrounded by a black circle represent the voxel with a higher \textit{grasp validness} than the threshold.
The three figures show that this classification is highly effective in rejecting those too-close or too-far grasps.
In addition, in Fig.~\ref{fig:volume_horizontal_8}, grasps at $x=0$, $y=\pm0.05$ region are rejected. Indeed, grasping this region is impossible due to the limitation of the hand's maximum aperture of 85 mm; thus, training data does not contain such grasps.
Similarly, in Fig.~\ref{fig:volume} grasps at $x=0.03$, $y=0.12$ are rejected.
This is also reasonable since the hand's base will collide with the objects in order to grasp this concave part.

\subsection{Simulation Benchmark}
\label{sec:simulation_experiment}

\begin{figure}[t]
    \centerline{\includegraphics[width=0.99\columnwidth]{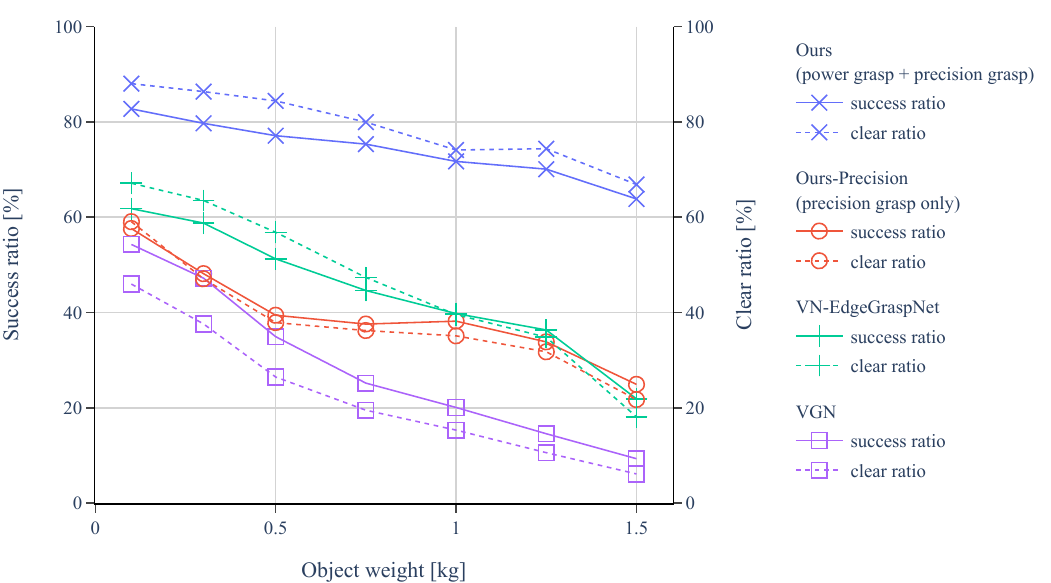}}
    \caption{Success ratio (SR) and clear ratio (CR) against multiple object weights.
    Our approach significantly outperformed the baselines, especially when the objects were heavy, showing the advantage of power grasp.
    }
    \label{fig:weight}
\end{figure}

In this subsection, we compare our approach (\textit{ours}) with two baselines: Breyer et al.~\cite{vgn} (\textit{VGN}) since we share the same backbone, and Huang et al.~\cite{edge_grasp} (\textit{VN-EdgeGraspNet}) since they reported state-of-the-art performance.
While both \cite{vgn} and \cite{edge_grasp} provide a pre-trained weight for Franka Emika's hand~\cite{panda}
\footnote[1]{\cite{edge_grasp} performed a real-robot validation with the Robotiq gripper in their paper, but they only made Franka Emika version's weight public.}
, we target Robotiq 2F-85~\cite{robotiq} hand; thus, we created a simulation model for both hands with an identical grasp force and contact parameters\footnote[2]{We set 40 N for grasp force, 0.75/0.5 for static/dynamic friction coefficient, 5 for Hunt Crossley dissipation, 10$^8$ for hydroelastic-modulus.}.
As the simulator, we used Drake~\cite{drake} since its hydroelastic contact model~\cite{hydroelastic_2019} allows for a stable simulation of power grasping by hands with a passive parallel-link mechanism.
To clarify the effect of power grasps, we also trained our model (\textit{Ours-Precision}) for Franka Emika's hand with only precision grasp annotation in the training data.

For the quantitative evaluation, we used the same metrics as \cite{vgn, s4g, regnet, vpn, edge_grasp, giga}: success ratio (SR) representing the number of successful grasps divided by the number of total trials, and clear ratio (CR) representing the number of successful grasps divided by the total number of objects.
Each scene is terminated after two consecutive grasp failures or detection failures.
We created 128 scenes with the mesh models provided by Breyer et al.~\cite{vgn}
\footnote[3]{
We merged the two test mesh sets for \textit{packed} and \textit{pile} and removed those that didn't fit our hand.}
which are used by \cite{giga, edge_grasp} also.

While SR and CR are intuitive metrics, they mix two distinct aspects of the grasp quality: positional accuracy and robustness against disturbance.
We, therefore, propose to evaluate grasp detectors against varying object weights since, in the lightweight region, they mainly capture positional accuracy, while in the heavy region, they capture both aspects.
The ability to support a heavier object also means a grasp can lift the same objects with less grasp force, which is critical for handling fragile objects.
Note that changing other physical properties, such as the friction coefficient, is another option, but it is harder to disentangle the two aspects.
For example, lower friction requires both high positional accuracy (because only highly antipodal grasps are acceptable in precision grasping) and high robustness (because the object will slip more easily).

Figure~\ref{fig:weight} plots the SR and CR against the object weight from 0.1 kg to 1.5 kg.
In the light-weight region (less than 0.5 kg), \textit{Ours-Precision} performed similarly with \textit{VGN} while \textit{VN-EdgeGraspNet} outperformed both.
This is reasonable because we share the same backbone as \textit{VGN}, and the author of \textit{VN-EdgeGraspNet} reported that their grasp representation is more advantageous.
In the heavy-weight region (more than 1 kg), however, \textit{Ours-Precision} outperformed \textit{VGN} and performed on par with \textit{VN-EdgeGraspNet}.
This indicates that even when limited to precision grasping, our \textit{gravity-rejection score} is beneficial in grasping heavy objects.
One hypothesis is that if the hand only supports precision grasping, introducing the \textit{gravigy-rejection score} to the edge-grasp representation will provide a better score.
Nevertheless, our focus is on leveraging power grasping.
\textit{Ours}, with all features including power grasp annotation, significantly outperformed the others, with 20.9 \% improvement for both SR and CR in the case of 0.1 kg and 41.9 / 48.7 \% improvement of SR / CR in the case of 1.5 kg compared with \textit{VN-EdgeGraspNet}.

It was surprising that even in the light-weight region, \textit{Ours} outperformed the other cases.
One reason is simply because the Robotiq 2F-85 hand has a slightly larger fingertip than that of the Franka Emika's hand, but we also observed that power grasping is more robust against position error.
In a power grasp, the palm or the proximal finger link hits the object first, then the hand closes, and the distal fingers wrap around it.
Even if the hand is inserted too deeply, the grasp is still valid as long as the object doesn't fall down in the case of a side grasp.
Conversely, even when the hand insertion is too shallow, the distal fingers' wrap-around motion still ``pulls" the object into the hand, or, in the worst case, results in a precision grasp.

Figure~\ref{fig:weight} also shows that \textit{Ours} had the minimum drop of SR / CR due to the increasing object weight.
This quantitatively proves the benefit of power grasping.
Qualitatively, we observed that power grasping frequently occurred for cylinder-shaped objects and that it was especially effective in lifting heavy cylinders laid down on the surface by a top-down grasp.
We also observed that when the hand attempts to perform a power grasp on a thin object laid down on the surface, the grasp typically fails due to the fingertip's collision with the surface and the finger's undesired displacement due to passive compliance.
However, such failure mode was rare, and in most cases, a precision grasp was selected.
This shows that utilizing the \textit{grasp validness} is effective in allowing the network to automatically fall back to precision grasping when power grasping is not available.

\subsection{Comparison on the Rotation Representation}
\label{sec:compare_rotation}
Following Qin et al.~\cite{s4g}, we also select the rotation matrix as the rotation representation due to its continuity in the SO(3)~\cite{rotation_matrix}.
However, \cite{s4g} lacks a discussion on how to select the basis vectors of the rotation matrix, i.e., how we should define the TCP coordinate.
A rotation matrix $R = [\bm{e}_x, \bm{e}_y, \bm{e}_z] \in \mathbb{R}^{3\times3} $ is constructed by three basis vectors.
The network estimates two of them $\tilde{\bm{e}}_x, \tilde{\bm{e}}_z$ and reconstructs $R$ by:
\begin{eqnarray}
    \mathbf{e}_x &=& \tilde{\mathbf{e}}_x / \vert \tilde{\mathbf{e}}_x\vert  \nonumber \\
    \mathbf{e}_z &=& \mathbf{e}^\prime_z / \vert \mathbf{e}^\prime_z\vert ,~\mathbf{e}^\prime_z = \tilde{\mathbf{e}}_z - (\mathbf{e}_x \cdot \tilde{\mathbf{e}}_z)\mathbf{e}_x \\
    \mathbf{e}_y &=& \mathbf{e}_z \times \mathbf{e}_x  \nonumber
    \label{eq:rotation}
\end{eqnarray}
which means it prioritizes $\tilde{\bm{e}}_x$ to decide two dimensions of the rotation and supplementarily uses $\tilde{\bm{e}}_z$ to decide the remaining one, while doesn't explicitly estimate the $\bm{e}_y$ direction.
We name this order as $\bm{e}_x$-$\bm{e}_z$ and compare it with the opposite order $\bm{e}_y$-$\bm{e}_z$, in addition to a quaternion version (see Fig.~\ref{fig:volume_experiment} left for the TCP coordinate definition.)
Table~\ref{tab:rotation} summarizes the result.
In addition to the large gap between the rotation matrix and quaternion, we also observe an improvement of $\bm{e}_x$-$\bm{e}_z$ over $\bm{e}_y$-$\bm{e}_z$.
This indicates that $\bm{e}_x$ is easier for the network to learn than $\bm{e}_y$.
In the other part of this paper, we learn $\bm{e}_x$ and $\bm{e}_z$, and use $\bm{e}_x$-$\bm{e}_z$ order for the inference.

\begin{table}[tbp]
    \caption{Success ratio (SR) and clear ratio (CR) under different rotation representation}
    \begin{center}
    \begin{tabular}{cccc}
    \hline
    & \makecell{Rotation Matrix \\ ($\mathbf{e}_x$-$\mathbf{e}_z$)} & \makecell{Rotation Matrix \\($\mathbf{e}_y$-$\mathbf{e}_z$)} & Quaternion \\
    \hline 
    SR [\%] & \textbf{71.7} & 68.0 & 57.8 \\
    CR [\%] & \textbf{74.1} & 69.2 & 61.0 \\
    \hline
    \end{tabular}
    \label{tab:rotation}
    \end{center}
\end{table}

\subsection{Validation with a Physical System}
\begin{figure}[tbp]
    \centerline{\includegraphics[width=0.9\columnwidth]{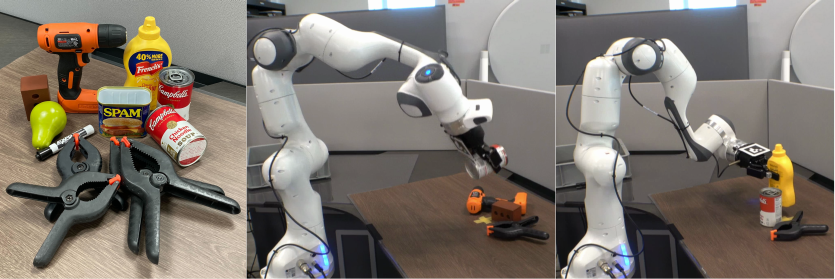}}
    \caption{
    Target objects (left) and capture (middle, right) of the validation.
    }
    \label{fig:panda}
\end{figure}

In this subsection, we validate our approach with a physical system, as shown in Fig.~\ref{fig:panda}.
Since we only have physical access to the Robotiq hand, we trained a baseline version of our model to mimic \cite{vgn} by (i) training data containing only precision grasps, (ii) we only used \textit{grasp validness} and ignored \textit{gravity-rejection score} for the inference, (iii) quaternion rotation representation.
We selected nine kinds of objects from the YCB dataset~\cite{ycb}, which we didn't use for training, and created 20 scenes with 51 objects in total in a simulator.
In \cite{vgn}, the objects were put in a box and dumped on the table.
However, since our target objects were often rigid and round-shaped, they frequently rolled out of the workspace.
Thus, we first created the scenes with the simulator, overlaid the scene's mesh with the real-time-streamed point cloud, and manually adjusted the objects' poses in the real world.
Depth image was acquired from a pair of RGB cameras and a learning-based stereo inference~\cite{learned_stereo}.
Table~\ref{tab:real_robot} summarizes the result.

The absolute number of SR / CR was lower than the value reported by \cite{vgn}.
We assume this is because of the different object properties.
In the case of \cite{vgn}, most objects were light and soft toys, which are relatively easy to grasp.
Indeed, our objects also contained such an object (\texttt{061\_foam\_brick}), and we observed that in any case, the grasp succeeds.
However, our objects contained more heavy and rigid objects.
For example, the unopened \texttt{005\_tomato\_soup\_can} weighed 350 g, and its hard metal exterior made the contact region small.
The object even harder to grasp was \texttt{035\_power\_drill}, with more than 600 g weight and unevenly distributed mass.
Other factors that may have contributed to the difference include the camera system, hand-eye calibration, and the underlying robot controller.
Similarly, due to the sim-to-real gap, we cannot compare the absolute SR / CR with the simulation evaluation in \ref{sec:simulation_experiment}.

In order to avoid such biases, our evaluation was performed under the exact same system configuration, and our approach significantly outperformed the baseline.
This result is consistent with the simulation and further justifies the effect of our contribution.

\begin{table}[tbp]
    \caption{Success ratio and clear ratio in the real robot validation}
    \begin{center}
    \begin{tabular}{cccccc}
    \hline
    & Objects & Trials & Success & SR [\%] & CR [\%] \\
    \hline
    Ours & 51 & 51 & 38 & \textbf{74.5} & \textbf{74.5} \\
    Baseline & 51 & 48 & 31 & 64.6 & 60.8 \\
    \hline
    \end{tabular}
    \label{tab:real_robot}
    \end{center}
\end{table}

\section{Limitation}
In this work, we reconstruct the hand pose through the location of the voxel corresponding to the TCP and the regressed rotation.
This representation, though, is fragile against the rotation regression error, since a small error in the rotation results in a large error in the contact regions.
We frequently observed failure cases in which the hand approach direction was inaccurate, causing the hand to grasp at an improper position (sometimes even not making contact with the object at all).
Another major failure case was when the finger collided with the object during the approaching motion due to an incorrect hand yaw rotation.
Our attempt in Sec.~\ref{sec:compare_rotation} is one countermeasure for such a problem, but there is still a large margin for improvement.
Contact-based grasp representations~\cite{contact_graspnet, edge_grasp} are more robust to rotation error.
Finally, it is difficult to generate power grasps due to the inconsistent contact region among different grasp modes, even though our \textit{gravity-rejection score} is also applicable to them.
An approach to detect power grasps through contact-based representations remains in our future work.

\section{Conclusion}
In this paper, we presented a data generation and learning framework to leverage power grasping in addition to precision grasping.
We proposed to train a neural network to predict a continuous \textit{gravity-rejection score} which is the magnitude of disturbance in the
gravity direction that a grasp can support. Unlike the grasp probability, this allows the network prioritize power grasping while still keeping precision grasping as a secondary choice.
We also proposed a data generation pipeline to efficiently create a dataset with \textit{gravity-rejection score} annotation.
We provided a qualitative analysis of the network output and showed the effect of \textit{gravity-rejection score} in generating power grasps.
For the quantitative evaluation, we proposed to evaluate the performance against varying object weights.
A simulation evaluation proved a significant improvement in our approach, where power grasping improves the grasp robustness to positioning accuracy error and gravity direction force disturbance.
Finally, we provided real robot validation, which showed that our approach is also effective in a physical system.

\bibliographystyle{IEEEtran}
\bibliography{citations.bib}

\vspace{12pt}

\end{document}